\documentclass[11pt,letterpaper]{article}
\usepackage{emnlp2017}
\usepackage{times}
\usepackage{latexsym}
\usepackage{amsmath}
\usepackage{bm}
\usepackage[utf8]{inputenc}
\usepackage{graphicx}
\usepackage{mathtools}
\usepackage{paralist}
\usepackage[labelformat=simple]{subfig}
\usepackage{booktabs}
\usepackage{multirow}
\usepackage{microtype}
\usepackage{tabularx}
\usepackage{hyperref}
\usepackage{xspace}
\usepackage{xcolor}
\usepackage{tablefootnote}
\usepackage{subfig}
\usepackage{capt-of}
\newcommand{\ie}{\textit{i.\,e.}\xspace}

\newcommand{\retrofit}{\textsc{Retrofit}\xspace}
\newcommand{\joint}{\textsc{Joint}\xspace}
\newcommand{\lstm}{\textsc{Lstm}\xspace}
\newcommand{\bilstm}{\textsc{BiLstm}\xspace}
\newcommand{\bow}{\textsc{Bow}\xspace}
\newcommand{\cnn}{\textsc{Cnn}\xspace}
\newcommand{\averaging}{\textsc{Ave}\xspace}
\newcommand{\rt}[1]{\rotatebox{90}{#1}}
\newcommand{\rrt}[1]{\rotatebox{45}{#1}}

\newcommand{\sd}[1]{\par \tiny #1}
\usepackage{scalefnt}

\emnlpfinalcopy

\def\emnlppaperid{27}

\title{Assessing State-of-the-Art Sentiment Models on\\
\hspace{46pt}  State-of-the-Art Sentiment Datasets}

\author{%
  Jeremy Barnes,
  Roman Klinger, \and
  Sabine Schulte im Walde \\
  Institut f\"ur Maschinelle Sprachverarbeitung\\ University of
  Stuttgart\\ Pfaffenwaldring 5b, 70569 Stuttgart, Germany\\
  \texttt{\{barnesjy,klinger,schulte\}@ims.uni-stuttgart.de}
}

\date{}

\begin{document}

\maketitle

\begin{abstract}
  There has been a good amount of progress in sentiment analysis over
  the past 10 years, including the proposal of new methods and the
  creation of benchmark datasets. In some papers, however, there is a
  tendency to compare models only on one or two datasets, either
  because of time restraints or because the model is tailored to a
  specific task. Accordingly, it is hard to understand how well a
  certain model generalizes across different tasks and datasets. In
  this paper, we contribute to this situation by comparing several
  models on six different benchmarks, which belong to different
  domains and additionally have different levels of granularity
  (binary, 3-class, 4-class and 5-class). We show that Bi-LSTMs
  perform well across datasets and that both LSTMs and Bi-LSTMs are
  particularly good at fine-grained sentiment tasks (\ie, with more
  than two classes). Incorporating sentiment information
  into word embeddings during training gives good results for datasets
  that are lexically similar to the training data. With our
  experiments, we contribute to a better understanding of the
  performance of different model architectures on different data
  sets. Consequently, we detect
  novel state-of-the-art results on the \textit{SenTube} datasets.
\end{abstract}

\section{Introduction}

The task of analyzing private states expressed by an author in text,
such as sentiment, emotion or affect, can give us access to a wealth
of hidden information to analyze product reviews \cite{Liu2005},
political views \cite{Speriosu2011}, or to identify potentially
dangerous activity on the Internet \cite{Forsyth2007}.  The first
approaches in this field of research depended on the use of words at a
symbolic level (unigrams, bigrams, bag-of-words features), where
generalizing to new words was difficult \cite{Pang2002,Riloff2003}.

Current state-of-the-art methods rely on features extracted in an
unsupervised manner, mainly through one of the existing pre-trained
word embedding approaches
\cite{Collobert2011a,Mikolov2013word2vec,Pennington2014}. These
approaches represent words as some function of their contexts,
enabling machine learning algorithms to generalize over tokens that
have similar representations, arguably giving them an advantage over
previous symbolic approaches.

In order to evaluate state-of-the-art models (both symbolic and
embedding-based), different datasets are used.  However, it is not
clear that a model that performs well on one certain dataset will
transfer well to other datasets with different properties. The work we
describe in this paper aims at discovering if there are certain models
that generally perform better or if there are certain models that are
better adapted to certain kinds of datasets. Ultimately, the goal of
this paper is to contribute to the current situation by supporting the
choice of a method for novel domains and datasets, based on
properties of the task at hand.

Our main contributions are, therefore, comparing seven approaches to
sentiment analysis on six benchmark datasets\footnote{The code and embeddings
for the best models are available at \url{http://www.ims.uni-stuttgart.de/data/sota_sentiment}}. We show that
\begin{compactitem}
\item bidirectional LSTMs perform well across datasets,
\item both LSTMs and bidirectional LSTMs are particularly good at
  fine-grained sentiment tasks,
\item and embeddings trained jointly for semantics and sentiment
  perform well on datasets that are similar to the training data.
\end{compactitem}

\section{Related Work}
\label{relatedwork}
This section discusses three approaches to sentiment analysis
and then describes in detail benchmark datasets which will be used
in the experiments.

\subsection{Approaches}
\label{approaches}

To analyze the performance of state-of-the-art methods across
datasets, we experiment with three approaches to sentiment analysis:
(1) updating pre-trained word embeddings using a neural classifier and
labeled data, (2) updating pre-trained word embeddings using a
semantic lexicon, and (3) training word embeddings to jointly maximize
a language model score and a sentiment
score. Sections~\ref{retrofitting}~to~\ref{supervised} discuss these
three approaches. We focus on sentiment-related methods,
however, where appropriate, we discuss general approaches which can be
adapted to this use case in a straight-forward manner as well.

\subsubsection{Retrofitting to Semantic Lexicons}
\label{retrofitting}

There have been several proposals to improve the quality of word
embeddings using semantic lexicons. \newcite{Yu2014} propose several
methods which combine the CBOW architecture \cite{Mikolov2013word2vec}
and a second objective function which attempts to maximize the
relations found within some semantic lexicon. They use both the
Paraphrase Database \cite{Ganitkevitch2013} and WordNet
\cite{Fellbaum99} and test their models on language modeling and
semantic similarity tasks. They report that their method leads to an
improvement on both tasks.

\newcite{Kiela2015} aim to improve embeddings by augmenting the
context of a given word while training a skip-gram model
\cite{Mikolov2013word2vec}. They sample extra context words, taken
either from a thesaurus or association data, and incorporate this into
the context of the word for each update. The evaluation is both
intrinsical, on word similarity and relatedness tasks, as well as
extrinsical on TOEFL synonym and document classification tasks. The
augmentation strategy improves the word vectors on all tasks.

\newcite{Faruqui2015} propose a method to refine word vectors by using
relational information from semantic lexicons (we will refer to this
method in this paper as \retrofit). They require a vocabulary $V =
\{w_{1},\ldots,w_{n}\}$, its word embeddings matrix $\hat{Q} =
\{\hat{q}_{1},\ldots,\hat{q}_{n}\}$, where each $\hat{q}_i$ is one
vector for one word $w_i$ and an
ontology $\Omega$, which they represent as an undirected graph $(V,E)$
with one vertex for each word type and edges $(w_{i}, w_{j}) \in E
\subseteq V \times V$. They attempt to
learn the matrix $Q = \{q_{1},\ldots,q_{n}\}$, such that $q_{i}$ is
similar to both $\hat{q}_{i}$ and $q_{j} \forall j$ for $(i,j)
\in E$.
Therefore, the objective function to minimize is
\[
\Psi(Q) = \sum_{i=1}^{n} \Bigg[ \alpha_{i}||q_{i} - \hat{q}_{i}||^{2} + \sum_{(i,j) \in E} \beta_{i,j} ||q_{i} - q_{j} ||^{2}   \Bigg]\,,
\]
where $\alpha$ and $\beta$ control the relative strengths of associations.

They use the XL version of the Paraphrase Database (PPDB-XL) dataset
\cite{Ganitkevitch2013}, which is a dataset of paraphrases as the
semantic lexicon, to improve the original vectors. This dataset
includes 8 million lexical paraphrases collected from bilingual
corpora, where words in language A are considered paraphrases if they
are consistently translated to the same word in language B. They then
test on the Stanford Sentiment Treebank \cite{Socher2013b}. They train
an L2-regularized logistic regression classifier on the average of the
word embeddings for a text and find improvements after retrofitting.

All above approaches show improvements over previous word embedding
approaches \cite{Mnih2012,Yu2014,Xu2014} on this data
set.

\subsubsection{Joint Training}
\label{joint}
\newcite{Maas2011} were the first to jointly train semantic and
sentiment word vectors. In order to capture semantic similarities,
they propose a probabilistic model using a continuous mixture model
over words, similar to Latent Dirichlet Allocation (LDA,
\citealp{Blei2003}).  To capture sentiment information, they include a
sentiment term which uses logistic regression to predict the sentiment
of a document. The full objective function is a combination of the
semantic and sentiment objectives. They test their model on several
sentiment and subjectivity benchmarks. Their results indicate that
including the sentiment information during training actually leads to
decreased performance.

\newcite{Tang2014} take the joint training approach and simultaneously
incorporate syntactic\footnote{We use the authors' terminology here,
  but make no assumptions that the distributional representation
  encodes information directly pertaining to syntax.} and sentiment
information into their word embeddings (we refer to this method as
\joint). They extend the word embedding approach of
\newcite{Collobert2011a}, who use a neural network to predict whether
an n-gram is a true n-gram or a ``corrupted'' version. They use the
hinge-loss
\begin{equation} \label{eq:1}
\begin{multlined}
\textrm{loss}_{\textrm{cw}}(t, t^{r}) =\\ \max(0,1 - f^{cw}(t) 
	                     + f^{cw}(t^{r}))
\end{multlined}
\end{equation}
and backpropagate the error to the corresponding word
embeddings. Here, $t$ is the original n-gram, $t^{r}$ is the corrupted
n-gram and $f^{cw}$ is the language model score.  \newcite{Tang2014}
add a sentiment hinge loss to the Collobert and Weston model, as 
\begin{equation}
\begin{multlined}
\textrm{loss}_{\textrm{s}}(t, t^{r}) =\\ \max(0, 1 - \delta_{s}(t) f^{s}_{1}(t) 
                          + \delta_{s}(t) f^{s}_{1}(t^{r}))\,,
\end{multlined}
\end{equation}
where $f^{s}_{1}$ is the predicted negative score and $\delta_{s}(t)$
is an indicator function that reflects the sentiment of a
sentence. $\delta_{s}(t)$ is $1$ if the true sentiment is positive and
$-1$ if it is negative. They then use a weighted sum of both scores to
create their sentiment embeddings:
\begin{equation}
\begin{multlined}
\textrm{loss}_{\textrm{combined}}(t,t^{r}) =\\ \alpha \cdot \textrm{loss}_{\textrm{cw}}(t,t^{r}) + 
                           (1 - \alpha) \cdot \textrm{loss}_{\textrm{s}}(t, t^{r})\,.
\end{multlined}
\end{equation}
This requires sentiment-annotated data for training both the syntactic
and sentiment losses, which they acquire by collecting tweets
associated with certain emoticons. In this way, they are able to
simultaneously incorporate sentiment and semantic information relevant
to their task. They test their approach on the SemEval 2013 twitter
dataset \cite{Nakov2013}, changing the task from three-class to binary
classification, and find that they outperform other approaches.

Overall, the joint approach shows promise for tasks with a large amount
of distantly-labeled data.

\subsubsection{Supervised training}
\label{supervised}
The most common approach to sentiment analysis is to use
pre-trained word embeddings in combination with a supervised
classifier. In this framework, the word embedding algorithm acts as a
feature extractor for classification.

Recurrent neural networks (RNNs), such as the \textsc{Long Short-Term
  Memory} network (\lstm) \cite{Hochreiter1997} or the
\textsc{Gated Recurrent Units} (GRUs) \cite{Chung2014}, are a variant
of a feed-forward network which includes a memory state capable of
learning long distance dependencies. In
various forms, they have proven useful for text classification tasks
\cite{Tai2015,Tang2016}. \newcite{Socher2013b} and \newcite{Tai2015}
use Glove vectors \cite{Pennington2014} in combination with a
recurrent neural networks and train on the Stanford Sentiment Treebank
\cite{Socher2013b}. Since this dataset is annotated for sentiment at
each node of a parse tree, they train and test on these annotated
phrases.

Both \newcite{Socher2013b} and \newcite{Tai2015} also propose various
RNNs which are able to take better advantage of the labeled nodes and
which achieve better results than standard RNNs. However, these models
require annotated parse trees, which are not necessarily available for
other datasets.

\textsc{Convolutional Neural Networks} (\cnn) have proven
effective for text classification \cite{Dossantos2014, Kim2014,
  Flekova2016}.  \newcite{Kim2014} use skipgram vectors
\cite{Mikolov2013word2vec} as input to a variety of Convolutional
Neural Networks and test on seven datasets, including
the Stanford Sentiment Treebank \cite{Socher2013b}. The best performing 
setup across datasets is a single
layer CNN which updates the original skipgram vectors during
training. 

Overall, these approaches currently achieve state-of-the-art results
on many datasets, but they have not been compared to retrofitting or
joint training approaches.

\subsection{Datasets}
\label{datasets}
We choose to evaluate the approaches presented in Section
\ref{approaches} on a number of different datasets from different
domains, which also have differing levels of granularity of class
labels. The Stanford Sentiment Treebank and SemEval 2013 shared-task
dataset have already been used as benchmarks for some of the
approaches mentioned in Section \ref{approaches}. Table
\ref{table:datamatrix} shows which approaches have been tested on
which datasets and Table~\ref{table:stats} gives an overview of the
statistics for each dataset.

\begin{table}[t]
\centering
\newcommand{\x}{+}
\newcommand{\m}{\textcolor{lightgray}{$-$}}
\begin{tabular}{lcccccccc}
\toprule
 	& \rt{\bow} & \rt{\averaging} & \rt{\retrofit} & \rt{\joint} & \rt{\lstm} & \rt{\bilstm} & \rt{\cnn} \\
\cmidrule(lr){1-1}\cmidrule(lr){2-2}\cmidrule(lr){3-3}\cmidrule(lr){4-4}\cmidrule(lr){5-5}\cmidrule(lr){6-6}\cmidrule(lr){7-7}\cmidrule(lr){8-8}
 \textit{SST-fine}   & \m &  \m  & \m & \m & \x & \x & \x \\
 \textit{SST-binary} & \m &  \x  & \x & \m & \x & \x & \x \\
 \textit{OpeNER}     & \x & \m & \m & \m & \m & \m & \m \\
 \textit{SenTube-A}  & \x & \m & \m & \m & \m & \m & \m  \\
 \textit{SenTube-T}  & \x & \m & \m & \m & \m & \m & \m \\
 \textit{SemEval}    & \m & \m & \m & \x & \m & \m & \m \\
\bottomrule
\end{tabular}
\caption{Mapping of previous state-of-the-art methods to previous
  evaluations on state-of-the-art data sets. An
  \x\ indicates that we are aware of a publication which reports on this combination and a \m\ indicates
  our assumption that no reported results are available.}
\label{table:datamatrix}
\end{table}

\newcommand{\smallsep}{\cmidrule(r){1-1}\cmidrule(r){2-2}\cmidrule(lr){3-3}\cmidrule(lr){4-4}\cmidrule(lr){5-5}\cmidrule(lr){6-6}\cmidrule(lr){7-7}}
\begin{table*}[ht]
\begin{center}
\begin{tabular}{lrrrccc}
\hline
                 & \multicolumn{1}{c}{Train} & \multicolumn{1}{c}{Dev.} & \multicolumn{1}{c}{Test}       & Number of Labels   &   Avg. Sentence Length &   Vocabulary Size \\
\smallsep
 \textit{SST-fine}        & 8,544 & 1,101 & 2,210 &            5 &             19.53 &          19,500 \\
 \textit{SST-binary}      & 6,920 & 872 & 1,821  &            2 &             19.67 &          17,539 \\
 \textit{OpeNER}          & 2,780 & 186 & 743   &            4 &              \phantom{0}4.28 &           \phantom{0}2,447 \\
 \textit{SenTube-A}  	     & 3,381 & 225 & 903   &            2 &             28.54 &          18,569 \\
 \textit{SenTube-T}		 & 4,997 & 333 & 1,334  &            2 &             28.73 &          20,276 \\
 \textit{SemEval}        & 6,021 & 890 & 2,376  &            3 &             22.40 &          21,163 \\
\hline
\end{tabular}
\caption{Statistics of datasets. Train, Dev., and Test refer to the number of examples for each subsection of a dataset. The number of labels corresponds to the annotation scheme, where: two is positive and negative; three is positive, neutral, negative; four is strong positive, positive, negative, strong negative; five is strong positive, positive, neutral, negative, strong negative.}
\label{table:stats}
\end{center}
\end{table*}

\subsubsection{Stanford Sentiment}
The Stanford Sentiment Treebank (\textit{SST-fine}) \cite{Socher2013b}
is a dataset of movie reviews which was annotated for 5 levels of
sentiment: strong negative, negative, neutral, positive, and strong
positive. It is annotated both at the clause level, where each node in
a binary tree is given a sentiment score, as well as at sentence
level. We use the standard split of 8544/1102/2210 for training,
validation and testing. In order to compare with
\newcite{Faruqui2015}, we also adapt the dataset to the task of binary
sentiment analysis, where strong negative and negative are mapped to
one label, and strong positive and positive are mapped to another
label, and the neutral examples are dropped. This leads to a slightly
different split of 6920/872/1821 (we refer to this dataset as
\textit{SST-binary}).

\subsubsection{OpeNER}
The \textit{OpeNER} dataset \citep{Agerri2013} is a dataset of hotel
reviews in which each review is annotated for opinions. An opinion
includes sentiment holders, targets, and phrases, of which only the
sentiment phrase is obligatory. Additionally, sentiment phrases are
annotated for four levels of sentiment: strong negative, negative,
positive and strong positive. We use a split of 2780/186/734
examples.

\subsubsection{Sentube Datasets}
The SenTube datasets \cite{Uryupina2014} are texts that are taken from
YouTube comments regarding automobiles and tablets. These comments are
normally directed towards a commercial or a video that contains
information about the product. We take only those comments that have
some polarity towards the target product in the video. For the
automobile dataset (\textit{SenTube-A}), this gives a 3381/225/903
training, validation, and test split. For the tablets dataset
(\textit{SenTube-T}) the splits are 4997/333/1334. These are annotated
for positive, negative, and neutral sentiment.

\subsubsection{Semeval 2013}
The SemEval 2013 Twitter dataset (\textit{SemEval}) \cite{Nakov2013}
is a dataset that contains tweets collected for the 2013 SemEval
shared task B. Each tweet was annotated for three levels of sentiment:
positive, negative, or neutral. There were originally 9684/1654/3813
tweets annotated, but when we downloaded the dataset, we were only
able to download 6021/890/2376 due to many of the tweets no longer
being available.

\section{Experimental Setup}
\label{experiment}

We compare seven approaches, five of which fall into the categories
mentioned in Section \ref{relatedwork}, as well as two baselines. The
models and parameters are described in Section \ref{models}. We test
these models on the benchmark datasets mentioned in Section
\ref{datasets}.

\subsection{Models}
\label{models}
\subsubsection{Baselines}

We compare our models against two baselines. First, we train an L2-regularized
logistic regression on a bag-of-words representation (\bow) of
the training examples, where each example is represented as a vector
of size $n$, with $n = |V|$ and $V$ the vocabulary. This is a standard
baseline for text classification.

Our second baseline is an L2-regularized logistic regression
classifier trained on the average of the word vectors in the training
example (\averaging). We train word embeddings using the
skip-gram with negative sampling algorithm \cite{Mikolov2013word2vec}
on a 2016 Wikipedia dump, using 50-, 100-, 200-, and 600-dimensional vectors,
a window size of 10, 5 negative samples, and we set the subsampling
parameter to $10^{-4}$. Additionally, we use the publicly available 300-dimensional
GoogleNews
vectors\footnote{\url{https://code.google.com/archive/p/word2vec/}} in
order to compare to previous work.

\subsubsection{Retrofitting}
We apply the approach by \newcite{Faruqui2015} and make use of the
code\footnote{\url{https://github.com/mfaruqui/retrofitting}}
released in combination
with the PPDB-XL lexicon, as this gave the best results for sentiment
analysis in their experiments. We train for 10 iterations. Following
the authors' setup, for testing we train an L2-regularized logistic
regression classifier on the average word embeddings for a phrase
(\retrofit).

\subsubsection{Joint Training}
For the joint method, we use the 50-dimensional sentiment embeddings
provided by \newcite{Tang2014}. Additionally, we create 100-, 200-,
and 300-dimensional embeddings using the code that is publicly
available\footnote{\url{http://ir.hit.edu.cn/~dytang}}. We use the
same hyperparameters as \newcite{Tang2014}: five million positive and
negative tweets crawled using hashtags as proxies for sentiment, a
20-dimensional hidden layer, and a window size of three. Following the authors'
setup, we concatenate the maximum, minimum and average vectors of the
word embeddings for each phrase. We then train a linear SVM on these
representations (\joint).

\subsubsection{Supervised Training}
We implement a standard \lstm which has an embedding layer that maps
the input to a \hbox{50-,} \hbox{100-,} \hbox{200-,} \hbox{300-,} or
600-dimensional vector, depending on the embeddings used to initialize
the layer.  These vectors then pass to an LSTM layer. We feed the
final hidden state to a standard fully-connected 50-dimensional dense
layer and then to a softmax layer, which gives us a probability
distribution over our classes. As a regularizer, we use a dropout
\cite{Srivastava2014} of 0.5 before the LSTM layer.

The \textsc{Bidirectional LSTM} (\bilstm) has the same architecture as
the normal LSTM, but includes an additional layer which runs from the
end of the text to the front. This approach has led to
state-of-the-art results for POS-tagging \cite{Plank2016}, dependency
parsing \cite{Kiperwasser2016} and text classification
\cite{Zhou2016}, among others. We use the same parameters as the LSTM,
but concatenate the two hidden layers before passing them to the dense
layer\footnote{For the neural models on the
  \textit{SST-fine} and \textit{SST-binary} datasets,
  we do not achieve results as
  high as \newcite{Tai2015} and \newcite{Kim2014}, 
  because we train our models only on
  sentence representations, not on the labeled phrase
  representations. We do this to be able to compare across
  datasets.}.

We also train a simple one-layer \cnn with one convolutional layer on
top of pre-trained word embeddings. The first layer is an embeddings
layer that maps the input of length \textit{n} (padded when needed) to
an $n\times R$ dimensional matrix, where $R$ is the dimensionality of the
word embeddings. The embedding matrix is then convoluted with filter
sizes of 2, 3, and 4, followed by a pooling layer of length 2. This is
then fed to a fully connected dense layer with ReLU
activations \cite{Nair2010} and finally to the softmax layer. We again use dropout
(0.5), this time before and after the convolutional layers.
  
For all neural models, we initialize our word representations with the
skip-gram algorithm with negative sampling
\cite{Mikolov2013word2vec}. For the 300-dimensional vectors, we use
the publicly available GoogleNews vectors. For the other dimensions
(50, 100, 200, 600), we create skip-gram vectors with a window size of 10,
5 negative samples and run 5 iterations. For out-of-vocabulary words,
we use vectors initialized randomly between -0.25 and 0.25 to
approximate the variance of the pre-trained vectors. We train our
models using ADAM \cite{Kingma2014a} and a minibatch size of 32 and
tune the hidden layer dimension and number of training epochs on the validation set.

\section{Results}
\begin{table*}[t]
\definecolor{blue}{cmyk}{0.2,0,0,0.1}
\newcommand{\hl}[1]{{\setlength{\fboxsep}{1pt}\colorbox{blue}{#1}}}
\newcommand{\best}[1]{\textbf{\setlength{\fboxsep}{1pt}\fbox{#1}}}
\newcommand{\sepb}{\cmidrule{1-1}\cmidrule(r){2-3}\cmidrule(lr){4-4}\cmidrule(lr){5-5}\cmidrule(lr){6-6}\cmidrule(lr){7-7}\cmidrule(lr){8-8}\cmidrule(lr){9-9}\cmidrule(lr){10-10}}
\newcommand{\sep}{\cmidrule(r){2-3}\cmidrule(lr){4-4}\cmidrule(lr){5-5}\cmidrule(lr){6-6}\cmidrule(lr){7-7}\cmidrule(lr){8-8}\cmidrule(lr){9-9}\cmidrule(lr){10-10}}
  \centering
  \renewcommand*{\arraystretch}{0.8}
  \setlength\tabcolsep{2.9mm}
  \newcolumntype{P}{>{\centering\arraybackslash}p{5mm}}
\scalefont{0.85}
  \begin{tabular}{lllllllllll}
    \toprule
    & \rrt{Model} & \rrt{Dim.}&\rrt{SST-fine} & \rrt{SST-binary} & \rrt{OpeNER} & \rrt{SenTube-A} & \rrt{SenTube-T} & \rrt{SemEval} & \rrt{Macro-Avg.} \\
    \sepb
    \multirow{6}{*}{\rt{Baselines}} & \multicolumn{2}{l}{\bow} & 40.3 & 80.7 & \hl{77.1\textsuperscript{4}}& \hl{60.6\textsuperscript{5}}& \hl{66.0\textsuperscript{5}}& 65.5& 65.0\\
    \sep
    \multirow{34}{*}{\rt{State-of-the-Art Methods}} & \multirow{4}{*}{\averaging} & 50 & 38.9 & 74.1 & 59.5& 62.0& 61.7& 58.1& 59.0 \\
    &                        & 100 & 39.7 & 76.7 & 67.2& 61.5& 61.8& 58.8& 60.9 \\
    &                        & 200 & 40.7 & 78.2 & 69.3& 60.6& 62.8& 61.1& 62.1 \\
    &                        & 300 & 41.6 & \hl{80.3}\textsuperscript{3} & 76.3& 61.5& 64.3& 63.6& 64.6 \\
    &                        & 600 & 40.6 & 79.1 & 77.0& 56.4& 62.9& 61.8& 63.0 \\
    \sepb
    & \multirow{4}{*}{\rt{\retrofit}} & 50 & 39.2 & 75.3 & 63.9& 60.6& 62.3& 58.1& 59.9 \\
    &                        & 100 & 39.7 & 76.7 & 70.0& 61.4& 62.8& 59.5& 61.7 \\
    &                        & 200 & 41.8 & 78.3 & 73.5& 60.0& 63.2& 61.2& 63.0 \\
    &                        & 300 & 42.2 & \hl{81.2}\textsuperscript{3} & 75.9& 61.7& 63.6& 61.8& 64.4 \\
    &                        & 600 & 42.9 & 81.1 & 78.3& 60.0& 65.5& 62.4& 65.0 \\
   \sep
   & \multirow{4}{*}{\rt{\joint}} & 50 & 35.8 & 70.6 & 72.9& 65.1& \best{68.1}& \hl{66.8}\textsuperscript{6}& 63.2 \\
   &                         & 100 & 34.3 & 70.8 & 67.0& 64.3& 66.4& 60.1& 60.5 \\
   &                         & 200 & 33.7 & 72.3 & 68.6& \best{66.2}& 66.6& 58.4& 61.0 \\
   &                         & 300 & 36.0 & 71.6 & 70.1& 64.7& 67.6& 60.8& 61.8 \\
   &                         & 600 & 36.9 & 74.0 & 75.8 & 63.7 & 64.2 & 60.9& 62.6 \\
   \sep
   &   \multirow{4}{*}{\rt{\lstm}} & 50 & 43.3 \sd{(1.0)} & 80.5 \sd{(0.4)}& 81.1 \sd{(0.4)}& 58.9 \sd{(0.8)}& 63.4 \sd{(3.1)}& 63.9 \sd{(1.7)}& 65.2 \sd{(1.2)}  \\ 
   &                         & 100 & 44.1 \sd{(0.8)} & 79.5 \sd{(0.6)}& 82.4 \sd{(0.5)}& 58.9 \sd{(1.1)}& 63.1 \sd{(0.4)}& 67.3 \sd{(1.1)}& 65.9 \sd{(0.7)}  \\
   &                         & 200 & 44.1 \sd{(1.6)} & 80.9 \sd{(0.6)}& 82.0 \sd{(0.6)}& 58.6 \sd{(0.6)}& 65.2 \sd{(1.6)}& 66.8 \sd{(1.3)}& 66.3 \sd{(1.1)}  \\
   &                         & 300 & \hl{45.3}\textsuperscript{1} \sd{(1.9)} & \hl{81.7}\textsuperscript{1}\sd{(0.7)}& 82.3 \sd{(0.6)}& 57.4 \sd{(1.3)}& 63.6 \sd{(0.7)}& 67.6 \sd{(0.6)}& 66.3 \sd{(1.0)}  \\
   & 							& 600 & 44.5 \sd{(1.4)} & \best{83.1} \sd{(0.9)}& 81.2 \sd{(0.8)}& 57.4 \sd{(1.1)}& 65.7 \sd{(1.2)}& 67.5 \sd{(0.7)}& 66.5 \sd{(1)}  \\
    \sep
    
   &    \multirow{4}{*}{\rt{\bilstm}} & 50 & 43.6 \sd{(1.2)} & 82.9 \sd{(0.7)}& 79.2 \sd{(0.8)}& 59.5 \sd{(1.1)}& 65.6 \sd{(1.2)}& 64.3 \sd{(1.2)}& 65.9 \sd{(1.0)}  \\ 
   &                         & 100 & 43.8 \sd{(1.1)} & 79.8 \sd{(1.0)}& 82.4 \sd{(0.6)}& 58.6 \sd{(0.8)}& 66.4 \sd{(1.4)}&65.2 \sd{(0.6)}& 66.0 \sd{(0.9)}  \\
   &                         & 200 & 44.0 \sd{(0.9)} & 80.1 \sd{(0.6)}& 81.7 \sd{(0.5)}& 58.9 \sd{(0.3)}& 63.3 \sd{(1.0)}& 66.4 \sd{(0.3)}& 65.7 \sd{(0.6)}  \\
   &                         & 300 & \hl{\best{45.6}}\textsuperscript{1} \sd{(1.6)} & \hl{82.6}\textsuperscript{1} \sd{(0.7)}& \best{82.5} \sd{(0.6)}& 59.3 \sd{(1.0)}& 66.2 \sd{(1.5)}& 65.1 \sd{(0.9)}& \best{66.9} \sd{(1.1)}  \\
   &							 & 600 & 43.2 \sd{(1.1)} & 83 \sd{(0.4)}& 81.5 \sd{(0.5)}& 59.2 \sd{(1.6)}& 66.4 \sd{(1.1)}& \best{68.5}\sd{(0.7)}& \best{66.9} \sd{(0.9)}  \\
    \sep
    
   &    \multirow{4}{*}{\rt{CNN}} & 50 & 39.9 \sd{(0.7)} & 81.7 \sd{(0.3)}& 80.0 \sd{(0.9)}& 55.2 \sd{(0.7)}& 57.4 \sd{(3.1)}& 65.7 \sd{(1.0)}& 63.3 \sd{(1.1)}  \\ 
   &                         & 100 & 40.1 \sd{(1.0)} & 81.6 \sd{(0.5)}& 79.5 \sd{(0.9)}& 56.0 \sd{(2.2)}& 61.5 \sd{(1.1)}& 64.2 \sd{(0.8)}& 63.8 \sd{(1.1)}  \\
   &                         & 200 & 39.1 \sd{(1.1)} & 80.7 \sd{(0.4)}& 79.8 \sd{(0.7)}& 56.3 \sd{(1.8)}& 64.1 \sd{(1.1)}& 65.3 \sd{(0.8)}& 64.2 \sd{(1.0)}  \\
   &                         & 300 & \hl{39.8}\textsuperscript{2} \sd{(0.7)} & \hl{81.3}\textsuperscript{2} \sd{(1.1)}& 80.3 \sd{(0.9)}& 57.3 \sd{(0.5)}& 62.1 \sd{(1.0)}& 63.5 \sd{(1.3)}& 64.0 \sd{(0.9)}  \\
   &							 & 600 & 40.7 \sd{(2.6)} & 82.7 \sd{(1.2)}& 79.2 \sd{(1.4)}& 56.6 \sd{(0.6)}& 61.3 \sd{(2)}& 65.9 \sd{(1.8)}& 64.4 \sd{(1.5)}  \\
  \bottomrule
  \end{tabular}
  \caption{Accuracy on the test sets. For all neural models we perform 5 runs and show the
    mean and standard deviation. The best results for each dataset is
    given in \best{bold} and results that have been previously
    reported are \hl{highlighted}. All results derive from our
    reimplementation of the methods. We describe significance
    values in the text and appendix. Footnotes refer to the work where a
    method was previously tested on a specific dataset, although not necessarily with the
    same results: [1] \newcite{Tai2015} [2] \newcite{Kim2014} [3]
    \newcite{Faruqui2015} [4] \newcite{Lambert2015} [5]
    \newcite{Uryupina2014} [6] \newcite{Tang2014}.}
  \label{results}
\end{table*}

\begin{figure*}
  \centering
  \includegraphics[width=1\textwidth]{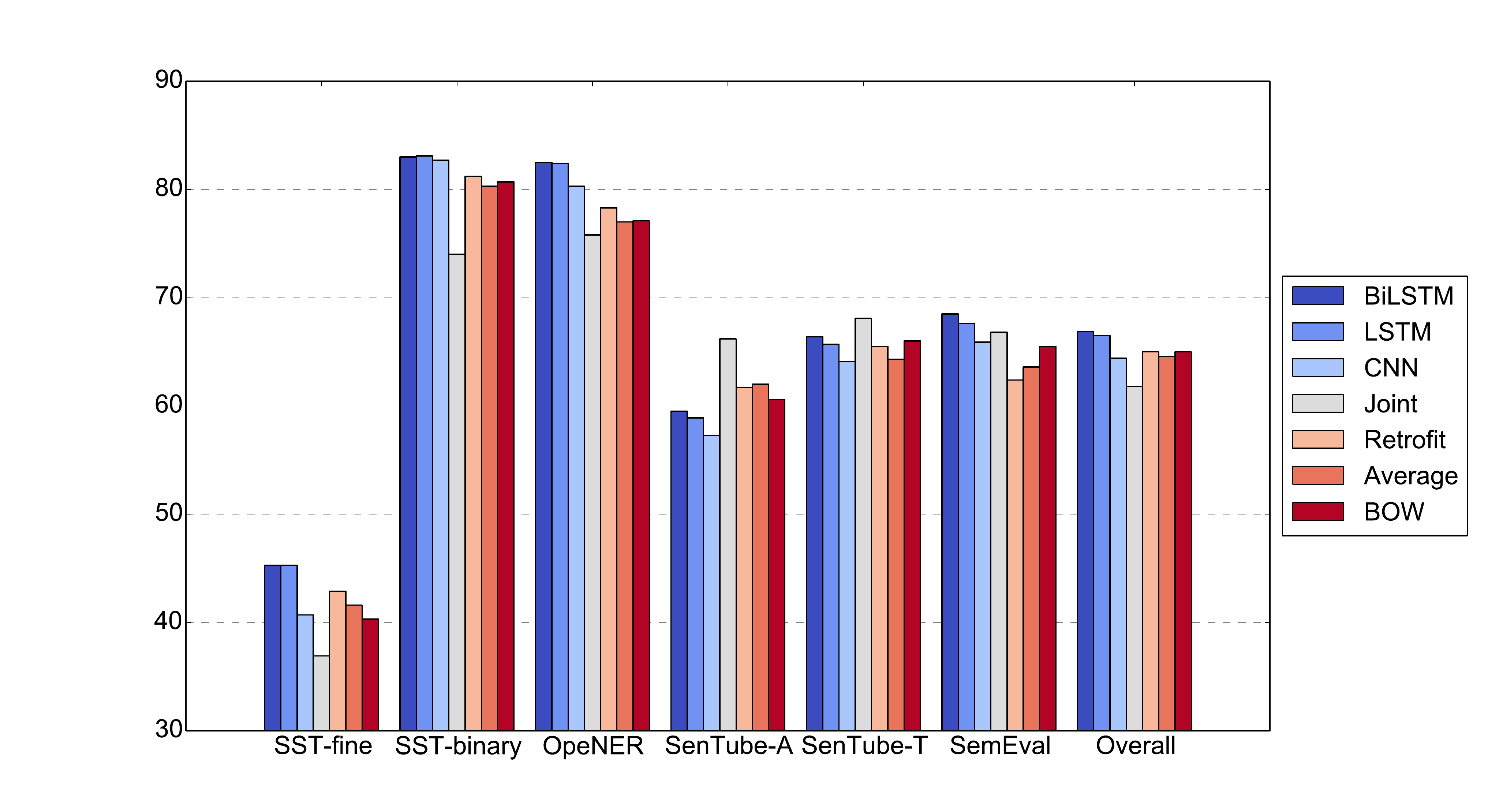}
  \caption{Maximum accuracy scores in percent for each model on the
    datasets. \lstm and \bilstm outperform other models on tasks with
    more than two labels (\textit{SST-fine}, \textit{OpeNER}, and
    \textit{SemEval}). \bow peforms well against more powerful
    models. \joint performs well on social media (\textit{SenTube-A},
    \textit{SenTube-T}, and \textit{SemEval}), but poorly on other
    tasks.}
  \label{fig:bar}
\end{figure*}

Table \ref{results} shows the results for the seven models across all
datasets, as well as the macro-averaged results. We visualize them in
Figure~\ref{fig:bar}. We performed random approximation tests
\cite{Yeh2000} using the \textit{sigf} package \cite{sigf06} with
10,000 iterations to determine the statistical significance of
differences between models. Since the reported accuracies for the
neural models are the means over five runs, we cannot use this
technique in a straightforward manner. Therefore, we perform the
random approximation tests between the runs\footnote{We compare the
  results from the first run of model A with the first run of model B,
  then the second from A with the second from B, an so forth. An
  alternative would have been to use a t-test, which is common in such
  setting. However, we opted against this as the independence
  assumptations for such test do not hold.  } and consider the models
statistically different if a majority (at least 3) of the runs are
statistically different ($p < 0.01$, which corresponds to $p < 0.05$
with Bonferroni correction for 5 hypotheses). The results of
statistical testing are summarized in Table~\ref{table:stats}.

Obviously, \textsc{BOW} continues to be a strong baseline: Though it
never provides the best result on a dataset, it gives better results
than \averaging on \textit{OpeNER}, \textit{SenTube-T}, and
\textit{SemEval}. Surprisingly, it also performs better than \joint on
the same sets except for \textit{SenTube-T}. Similarly, it outperforms
\retrofit on \textit{SenTube-T} and \textit{SemEval}.

\retrofit performs better than \cnn on \textit{SST-fine} and \joint on
\textit{SST-fine}, \textit{SST-binary}, and \textit{OpeNER}.  It also
improves the results of \averaging across all datasets but
\textit{SenTube-A} and \textit{SemEval} datasets.

Although \joint does not perform well across datasets and, in fact,
does not surpass the baselines on some
datasets, it does lead to good results on
\textit{SemEval} and to 
state-of-the-art results on \textit{SenTube-A} and \textit{SenTube-T}.

Similarly to \retrofit, \cnn does not outperform any of the other
methods on any dataset. As said, this method does not beat
the baseline on \textit{SST-fine}, \textit{SenTube-A}, and
\textit{SenTube-T}. However, it outperforms the
\averaging baseline on \textit{SST-binary} and \textit{OpeNER}.

The best models are \lstm and \bilstm. The best overall model is
\textsc{BiLSTM}, which outperforms the other models on half of the
tasks (\textit{SST-fine}, \textit{Opener}, and \textit{SemEval})
and consistently beats the baseline. This is in line with other
research \cite{Plank2016,Kiperwasser2016,Zhou2016}, which suggests
that this model is very robust across tasks as well as datasets. 
The differences in performance between \lstm
and \bilstm, however, are only significant
(p $< 0.01$) on the \textit{SemEval} dataset.

We also see that the difference in performance between the two LSTM
models and the others is larger on datasets with fine-grained labels
(\bilstm 45.6 and \lstm 45.3 vs.\ an average of 40 for all others on
the \textit{SST-fine} and \bilstm 83 and \lstm 83.1 vs.\ an average
of 76.5 on \textit{OpeNER}). These differences between the LSTM models
and other models are statistically significant, except for the difference
between \bilstm and \cnn at 50 dimensions on the \textit{OpeNER} dataset.

Our analysis of different dimensionalities as input for the classification
models reveals that, typically, the higher dimensional vectors (300 or 600)
outperform lower dimensions. The only differences are in \joint for 
\textit{SenTube-T} and \textit{SemEval} and \lstm for \textit{SenTube-A}
and \averaging on all datasets except \textit{OpeNER}.

\begin{figure*}
     \centering
\includegraphics[width=\textwidth]{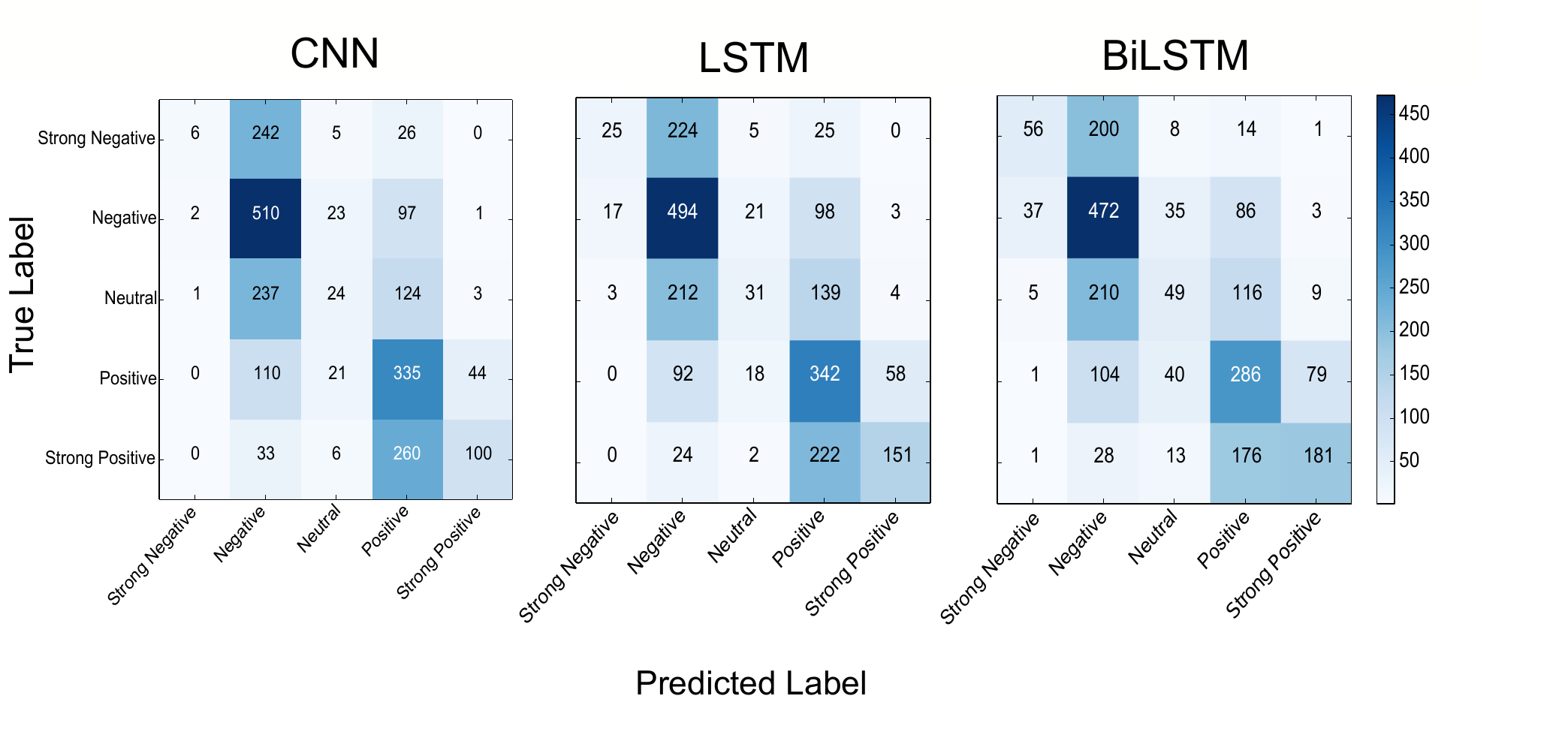}
     \caption{Confusion matrices of \cnn, \lstm, and \bilstm on
       \textit{SST-fine} dataset. We can see that both \lstm and
       \bilstm perform much better than \cnn on strong negative,
       neutral, and strong positive classes.}
     \label{fig:confmatrix}
\end{figure*}

\section{Discussion}

While approaches that average the word embeddings for a sentence
are comparable to state-of-the-art results \cite{Iyyer2015}, \averaging
and \retrofit do not perform particularly well. This is likely due to 
the fact that logistic regression lacks the non-linearities which \newcite{Iyyer2015}
found helped, especially at deeper layers. Averaging all of the embeddings
for longer phrases also seems to lead to representations that do not
contain enough information for the classifier.

We also experimented with using large sentiment
lexicons as the semantic lexicon for retrofitting, but found that this
hurt the representation more than it helped. We believe this is
because there are not enough kinds of relationships to exploit the
graph structure and by trying to collapse all words towards either a
positive or negative center, too much information is lost.

We expected that \joint would perform well on \textit{SemEval}, given
that it was designed for this task, but it was surprising that it
performed so well on the \textit{SenTube} datasets. It might be due to
the fact that comments for these three datasets are comparably
informal and make use of emoticons and Internet jargon. We performed
a short analysis of datasets (shown in Table
\ref{emoticonstats}), where we take frequency of
emoticons usage as an indirect indicator of informal speech and found
that, indeed, the frequency of emoticons in the \textit{SemEval} and
\textit{SenTube} datasets diverges significantly from the other
datasets. 
The fact that \joint is distantly trained on 
similar data gives it an advantage over other models on these
datasets. This
leads us to believe that this approach would transfer well to novel
sentiment analysis tasks with similar properties.

\begin{table}[tb]
\centering
\begin{tabular}{lrr}
\toprule
 $\chi^{2}$ with SemEval   & \multicolumn{1}{c}{$\chi^{2}$} & \multicolumn{1}{c}{ p-value} \\
\cmidrule(r){1-1}\cmidrule(rl){2-2}\cmidrule(l){3-3}
 SST-fine            & 19.408 &     0.002 \\
 SST-binary          & 19.408 &     0.002 \\
 OpeNER              & 19.408 &     0.002 \\
 SenTube-A           &  9.305 &     0.097 \\
 SenTube-T           &  7.377 &     0.194 \\
\hline
\end{tabular}
\caption{$\chi^{2}$ statistics comparing the frequency of the
  following emoticons over the different datasets,  :), :(, :-), :-(,
  :D, =). The difference in frequency of emoticons between the SemEval
  and  SenTube datasets is not significant (p $> 0.05$), while for SST
  and OpeNER it is (p $< 0.05$).}
\label{emoticonstats}
\end{table}

The fact that \cnn performs much better on \textit{OpeNER} may be due
to the smaller size of the phrases (an average of 4.28 vs.\ 20+ for other
datasets), however, further analyses to prove this are needed.
 
The good results that both LSTM models achieved on the more
fine-grained sentiment datasets (\textit{SST-fine} and
\textit{OpeNER}) seem to indicate that LSTMs are able to learn
dependencies that help to differentiate strong and weak versions of
sentiment better than other models. This is supported by the confusion
matrices shown in Figure~\ref{fig:confmatrix}. This makes them natural
candidates for fine-grained sentiment analysis tasks.

\lstm perfoms better than \bilstm on two datasets but these differences are
not statistically significant. 

The effect of the dimensionality of the input for the classification
models suggests that larger dimensionalities tend to perform better.
This seems particularly true for \retrofit, which continues gaining 
performance even at 600 dimensions. Most other approaches perform slightly 
better at 600 dimensions, but \averaging consistently performs worse at
600 than at 300.

\section{Conclusions}
The goal of this paper has been to discover which models perform
better across different datasets. We compared state-of-the-art models
(both symbolic and embedding-based) on six benchmark datasets with
different characteristics and showed that Bi-LSTMs perform well across
datasets and that both \textsc{LSTMs} and \text{Bi-LSTMs} are
particularly good at fine-grained sentiment tasks. Additionally,
incorporating sentiment information into word embeddings during
training gives good results for datasets that are lexically similar to
the training data. Finally, we reported a new state of the art on
the \textit{SenTube} datasets.
	  
\label{conclusion}

\section*{Acknowledgments}
We thank Sebastian Pad\'o for fruitful discussions. Thanks to Diego
Frassinelli for help with the statistical tests. This work has been
partially supported by the DFG Collaborative Research Centre SFB 732.

\begin{figure*}[t]
\small
\renewcommand*{\arraystretch}{0.5}
\setlength\tabcolsep{2.5mm}
\newcommand{\m}{\textcolor{black}{}}
\newcommand{\sst}{\textit{SST-fine}}
\newcommand{\sstt}{\textit{SST-binary}}
\newcommand{\op}{\textit{OpeNER}}
\newcommand{\sa}{\textit{SenTube-A}}
\newcommand{\st}{\textit{SenTube-T}}
\newcommand{\sem}{\textit{SemEval}}
\newcommand{\sep}{\cmidrule(lr){1-1}\cmidrule(lr){2-2}\cmidrule(lr){3-3}\cmidrule(lr){4-4}\cmidrule(lr){5-5}\cmidrule(lr){6-6}\cmidrule(lr){7-7}\cmidrule(lr){8-8}}
\newcolumntype{Y}{>{\centering\arraybackslash}X}

\begin{tabularx}{\textwidth}{lYYYYYYY}
\toprule
 	& \bow & \averaging & \retrofit & \joint & \lstm & \bilstm & \cnn \\
\sep

\multirow{6}{*}{\bow} & 	\m	& \sst & \sst  & \sst  & \sst  & \sst  & \sstt \\
 					  & 	\m	& \sa  & \op   & \sstt & \sstt & \sstt & \op \\
 					  & 	\m	& \st  & \sa   & \op   & \op   & \op   & \sa \\
 					  & 	\m  & \sem & \sem  & \sa   & \sem  & \sem  & \st \\
 					  & \m  &      &       & \st   & 	  &       & \\
 					  & \m  &      &       & \sem  &       &       & \\ 
 \sep

\multirow{6}{*}{\averaging} & 		& \m   & \sst  & \sst  & \sst  & \sst  & \sst \\
 					  & 		& \m   & \sstt & \sstt & \sstt & \sstt & \sstt \\
 					  & 	3	& \m   & \sa   & \op   & \op   & \op   & \op \\
 					  & 	    & \m   & \st   & \sa   & \sa   & \sa   & \sa \\
 					  &     & \m   &       & \st   & \st   & \sem  & \sem \\
 					  &     & \m   &       & \sem  & \sem  &       & \\
 \sep

\multirow{6}{*}{\retrofit}    & 		& 	   & \m  & \sst  & \sst    & \sst  & \sst \\
 					 		 & 		& 	   & \m	 & \sstt & \sstt   & \sstt & \sstt \\
 					 		 & 	3	& 	3  & \m   & \op   & \op    & \op   & \sa \\
 					    		 & 	    & 	   & \m   & \sa   & \sa    & \sa   & \sem \\
 					 		 &     & 	   &  \m   & \st   & \sem  & \sem  &  \\
 					 		 &     &		   &  \m   & \sem  & 	   &       & \\
 \sep 
 
\multirow{6}{*}{\joint}      & 		& 	   &       & \m  & \sst    & \sst  & \sst \\
 					 		 & 		& 	   & 	   & \m & \sstt   & \sstt & \sstt \\
 					 		 & 	3	& 	3  & 3     & \m   & \op     & \op   & \op \\
 					    		 & 	    & 	   &       & \m   & \sa     & \sa   & \sa \\
 					 		 &     & 	   &       & \m   & \st     & \st  & \st \\
 					 		 &     &		   &       &       &     	&  \sem     & \sem\\
 \sep 

\multirow{6}{*}{\lstm   }    & 		& 	   &      & 		  & \m    & \sem  & \sst \\
 					 		 & 		& 	   & 	  & 		  & \m    & 		  & \op \\
 					 		 & 	4	& 	5  & 4    & 3   	  & \m    & 		  & \sa \\
 					    		 & 	    & 	   &      & 		  & \m    & 		  & \st \\
 					 		 &     & 	   &      & 		  & \m  	  & 		  &  \sem \\
 					 		 &     &		   &      & 		  & \m    &        & \\
 \sep 

\multirow{6}{*}{\bilstm}    & 		& 	   &     &      &        & \m  & \sst \\
 					 		 & 		& 	   & 	 &      &        & \m & \op \\
 					 		 & 	4	& 	5  & 5   & 4    & 1		 & \m   & \sa \\
 					    		 & 	    & 	   &     &      &        & \m   & \st \\
 					 		 &     & 	   &     &      &         & \m  & \sem \\
 					 		 &     &		   &     &      & 	     &    \m   & \\
 \sep 

\multirow{6}{*}{\cnn}    & 		& 	   &     &       &         &       & \m \\
 					 		 & 		& 	   &  	 &       &         &       & \m \\
 					 		 & 	2	& 	3  & 2   & 3     & 0       & 0     & \m \\
 					    		 & 	    & 	   &     &       &         &       & \m \\
 					 		 &     & 	   &     &       &         &       & \m \\
 					 		 &     &		   &     &       & 	       &       & \m \\
 
\bottomrule
\end{tabularx}
\caption{Results of the statistical analysis described in Section
  \ref{results} for the best performing dimension of embeddings, where
  applicable. Datasets where there is a statistical difference (above
  diagonal) and number of datasets where a model on the Y axis is
  statistically better than a model on the X axis (below diagonal).}
\label{table:stats}
\end{figure*}

\bibliography{lit}
\bibliographystyle{emnlp_natbib}

\end{document}